\documentclass[final,5p,times,twocolumn,authoryear]{elsarticle}
\usepackage{amsmath}

\usepackage{amssymb}
\usepackage{wrapfig}
\usepackage{float}
\usepackage{hyperref}
\usepackage{amsmath}
\usepackage{multirow} 
\usepackage{url}
\usepackage{xcolor}
\usepackage{microtype}
\makeatletter
\renewcommand\paragraph{\@startsection{paragraph}{4}{\z@}%
  {3.25ex \@plus1ex \@minus.2ex}%
  {-1em}%
  {\normalfont\normalsize\itshape}}
\makeatother

\journal{-}

\begin{document}
\newcommand{\networkName}{INSPIRE-GNN}
\begin{frontmatter}

\title{INSPIRE-GNN: Intelligent Sensor Placement to Improve Sparse Bicycling Network Prediction via Reinforcement Learning Boosted Graph Neural Networks}

\author[1]{Mohit Gupta\corref{cor1}}
\cortext[cor1]{Corresponding author Email: \href{mailto:mohit.gupta1@monash.edu}{mohit.gupta1@monash.edu} (M. Gupta)}



\author[1]{Debjit Bhowmick}
\author[2]{Rhys Newbury}
\author[3]{Meead Saberi}
\author[4]{Shirui Pan}
\author[1]{Ben Beck}

\address[1]{School of Public Health and Preventive Medicine, Monash University, Melbourne, Australia}
\address[2]{Department of Electrical and Computer System Engineering, Monash University, Melbourne, Australia}
\address[3]{School of Civil and Environmental Engineering, UNSW, Sydney, Australia}
\address[4]{School of Information and Communication Technology, Griffith University, Brisbane, Australia}


\begin{abstract}
Accurate link-level bicycling volume estimation is essential for sustainable urban transportation planning. 
However, many cities face significant challenges of high data sparsity due to limited bicycling count sensor coverage. 
To address this issue, we propose \networkName, a novel Reinforcement Learning (RL)-boosted hybrid Graph Neural Network (GNN) framework designed to optimize sensor placement and improve link-level bicycling volume estimation in data-sparse environments. 
\networkName{} integrates Graph Convolutional Networks (GCN) and Graph Attention Networks (GAT) with a Deep Q-Network (DQN)-based RL agent, enabling a data-driven strategic selection of sensor locations to maximize estimation performance. 
Applied to Melbourne’s bicycling network, comprising 15,933 road segments with sensor coverage on only 141 road segments (99\% sparsity) — \networkName{} demonstrates significant improvements in volume estimation by strategically selecting additional sensor locations in deployments of 50, 100, 200, and 500 sensors. 
Our framework outperforms traditional heuristic methods for sensor placement such as betweenness centrality, closeness centrality, observed bicycling activity and random placement, across key metrics such as Mean Squared Error (MSE), Root Mean Squared Error (RMSE) and Mean Absolute Error (MAE). 
Furthermore, our experiments benchmark \networkName{} against standard machine learning and deep learning models in the bicycle volume estimation performance, underscoring its effectiveness. 
Our proposed framework provides transport planners actionable insights to effectively expand sensor networks, optimize sensor placement and maximize volume estimation accuracy and reliability of bicycling data for informed transportation planning decisions.
 
\end{abstract}

\begin{keyword}
reinforcement learning, graph neural networks, deep learning, bicycle volume estimation, machine learning, data sparsity
\end{keyword}

\end{frontmatter}


\section{Introduction}
Urban bicycling networks have gained significant attention worldwide as cities aim to promote bicycling as an active and sustainable mode of transportation. 
Bicycling offers multiple benefits such as improved public health, reduced air and noise pollution, decreased traffic congestion and greenhouse gas emissions \citep{de2007determining, wen2008inverse, de2010health, lindsay2011moving, grabow2012air}.
Many cities are investing in developing or upgrading their bicycling infrastructure such as protected and painted bike lanes, dedicated bike paths and bike bridges to encourage greater uptake of bicycling as a sustainable mode of transport \citep{buehler2012cycling}. 
However, the importance of bicycling volume estimation extends beyond just infrastructure investment. 
Accurate link-level estimates are essential for evaluating the effectiveness of infrastructure, policy and behavioral interventions aimed at promoting bicycling \citep{cooper2017using}. 
Detailed bicycling volume data enable decision makers to identify bicycling patterns, assess baseline demand and evaluate the outcomes of the interventions \citep{ATAP2022}. 
Such data are also critical for safety assessments, providing denominators to calculate crash risk per bicycle trip \citep{reynolds2009impact} and supporting equitable investment decisions \citep{winters2017cycling}.
Therefore, link-level volume estimation forms the foundation for network-wide monitoring, evidence-based planning, performance evaluation and strategic scaling of bicycling initiatives.

\begin{figure}
    \centering
    \includegraphics[width=1\linewidth]{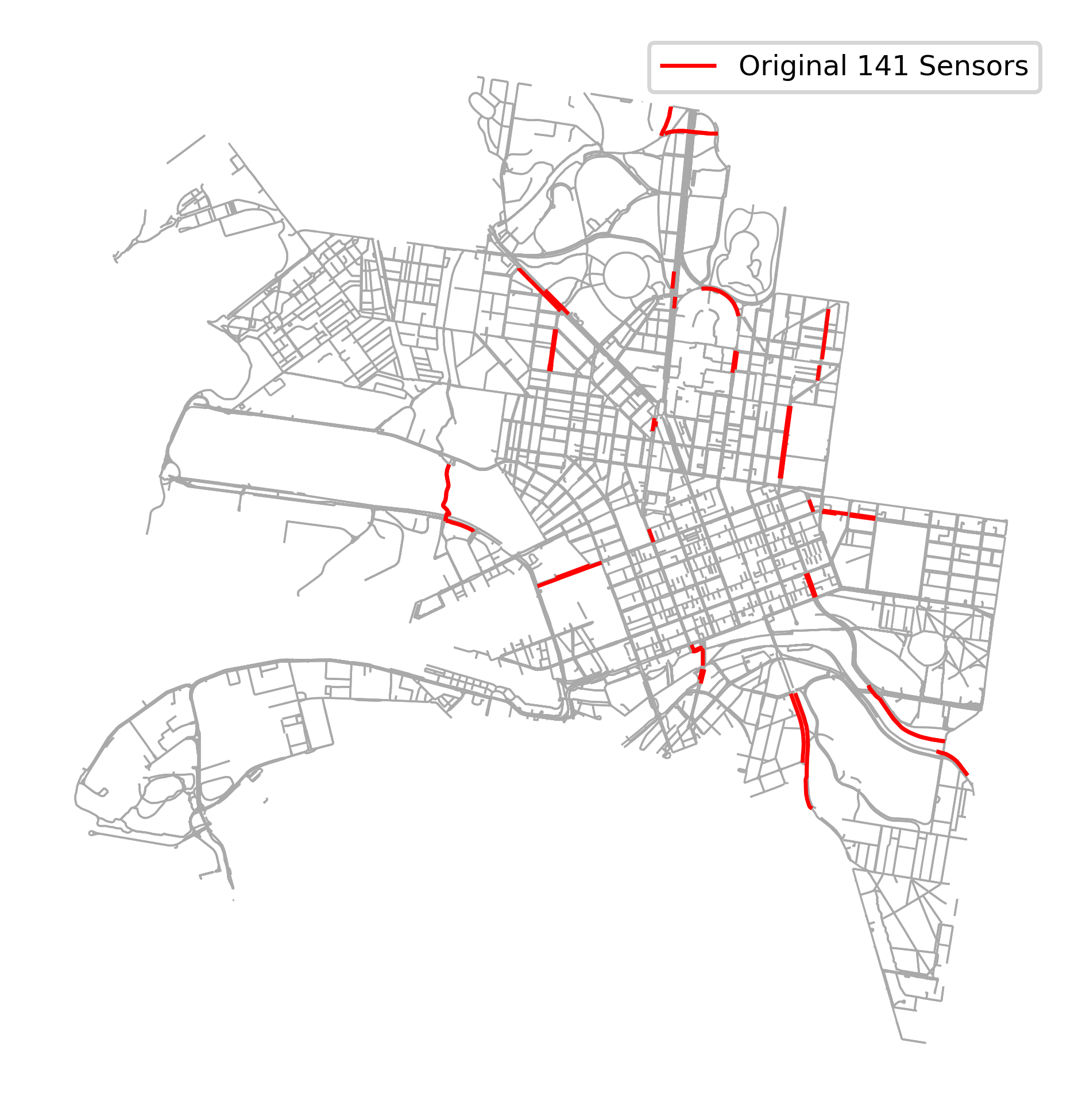}
    \caption{Spatial distribution of the existing 141 bicycle traffic sensors across Melbourne's road network. The sparse sensor coverage (approximately 1\% of total segments) highlights the challenge of data sparsity in bicycling volume estimation.}
    \label{fig:original_sensor_placement}
\end{figure}

A critical barrier to achieve accurate volume estimation is the extreme sparsity of bicycling data, as most cities lack sufficient sensor coverage required to monitor their sprawling bicycling network \citep{bhowmick2023systematic}. 
For instance as shown in Fig.~\ref{fig:original_sensor_placement} - in Melbourne, Australia, the bicycling road network comprises approximately 15,933 road/path segments, yet only 141 of these segments ($<1\%$) are equipped with sensors to record bicycling counts. This results in extreme (99\%) data sparsity, posing substantial challenges for accurate link-level volume estimation \citep{winters2016bike}. 
Unlike motorized transportation networks, which benefit from extensive sensor coverage using loop detectors, cameras, and automated traffic counters \citep{gu2017traffic, klein2006traffic}, bicycling networks have historically been overlooked, leading to significant gaps in sensor coverage and high data sparsity \citep{buehler2016bikeway}. 
These challenges impede the ability to accurately estimate and analyze bicycling patterns.

Traditional approaches to bicycling volume estimation such as statistical models and machine learning methods, rely on datasets with sufficient coverage and representation to generalize effectively \citep{griswold2011pilot, lin2020modeling, livingston2021predicting, bhowmick2023systematic}. However, these methods struggle in sparse data environments as the limited availability of labeled data severely constrains their predictive capabilities \citep{salon2016estimating}. 
They often fail to capture the complex spatial interdependencies inherent to transportation networks, further limiting their ability to accurately model bicycling volumes across urban road networks. Furthermore, existing Direct Demand models often use data from a small number of sensor locations which are typically placed in areas of high bicycling activity \citep{jestico2016mapping, kwigizile2019integrating, bhowmick2023systematic}. As a result, these models are biased and fail to generalize effectively when applied to street networks with low or no bicycle counts which undermines their reliability for network-wide estimation.

While recent advances in deep learning, particularly Graph Neural Networks (GNNs) have been effective for similar challenges in motorized traffic prediction by exploiting the connectivity and relational structure of road networks \citep{li2017diffusion, yu2017spatio, zhang2018gaan, zhao2019t, wu2020comprehensive}, their application in bicycling network is constrained by the challenge of extreme data sparsity.

To address these challenges, we build upon our prior work using Graph Convolutional Networks (GCNs) to estimate link-level bicycling volumes in City of Melbourne, which demonstrated strong performance up to moderate (80\%) data sparsity but faced limitations under extreme conditions \citep{gupta2024evaluating}. 
In this paper, we introduce \networkName{} (\textbf{In}telligent \textbf{S}ensor \textbf{P}lacement to \textbf{I}mprove Sparse bicycling Network Prediction via \textbf{Re}inforcement Learning-Enhanced Graph Neural Networks), a novel framework designed to optimize sensor placement and improve link-level bicycling volume estimation in data sparse networks. \networkName{} integrates two advances: 

\begin{itemize}
    \item \textbf{Hybrid-GNN Architecture:} INSPIRE-GNN integrates \textbf{Graph Convolutional Networks (GCNs)} and \textbf{Graph Attention Networks (GATs)} to capture both local road segment characteristics and global network topology, enabling robust predictions even in data sparsity.
    
    \item \textbf{Reinforcement Learning-Driven Policy Learning:}  A \textbf{Deep Q-Network (DQN)} agent is trained to sequentially select the most informative sensor locations that maximize information gain. By framing sensor placement as a sequential decision-making problem, our RL module learns an optimal deployment policy that significantly reduces predictive uncertainty across the entire network and yield the most significant improvement in the link-level bicycling volume estimation performance of the Hybrid-GNN Architecture. 
\end{itemize}

We apply \networkName{} to the bicycling network in Melbourne, utilizing crowd-sourced bicycle volume data from Strava \citep{StravaMetro} and infrastructure data from OpenStreetMap (OSM) \citep{OpenStreetMap}. 
To evaluate its effectiveness, we systematically simulate additional sensor deployments incrementally by adding 50, 100, 200, and 500 additional sensors and compare INSPIRE-GNN’s RL sensor placement policy with traditional heuristics such as random selection, betweenness centrality, closeness centrality, and observed bicycling activity \citep{freeman1977set, newman2005measure, paluch2020optimizing, senturk2014connectivity}. 
Furthermore, we benchmark the bicycling volume estimation performance of our hybrid GNN model against standard statistical and machine learning methods including decision tree, random forest, gradient boosting, multi-layer perceptron (MLP), and convolutional neural network (CNN) — to demonstrate its superiority under various sensor deployment scenarios.

Our primary contributions are as follows:
\begin{itemize}
    \item \textbf{Develop an advanced framework:} We introduce INSPIRE-GNN, an RL-enhanced hybrid GNN framework tailored for optimal sensor placement in data-sparse bicycling networks.
    \item \textbf{Improve link-level volume estimation:} The framework significantly enhances the accuracy of link-level bicycling volume predictions through strategic sensor placement.
    \item \textbf{Benchmark performance:} We compare INSPIRE-GNN’s RL sensor placement strategy against heuristic sensor placement strategies such as (sensor placement link selection based on) random selection, observed cycling activity, betweenness and closeness centrality and compare the volume estimation performance of INSPIRE-GNN’s hybrid (GCN + GAT) with standard machine learning and deep learning models.
    \item \textbf{Perform ablation studies:} We quantify the contribution of each component of \networkName{} (i.e. RL module, GCN, GAT) to the overall performance of the framework.
    \item \textbf{Provide practical insights:} Our results offer actionable recommendations for sensor network expansion, empowering urban planners to make informed decisions.
\end{itemize}

By effectively combining advanced graph-based learning with an intelligent RL-based sensor placement strategy, INSPIRE-GNN provides a data-driven framework for improving the accuracy of bicycling volume estimation in urban networks with sparse sensor coverage and high data sparsity. 
The findings have practical implications for cities aiming to expand their sensor networks strategically, ensuring that limited resources are utilized optimally for the greatest impact.

\section{Literature Review} \label{sec:litreview}
\subsection{Link-Level Bicycle Volume Estimation}
Accurate estimation of bicycling volumes has been a persistent challenge in urban transportation research \citep{bhowmick2023systematic}. 
Traditional approaches often rely on direct counts from temporary sensors, household travel surveys, or manual counts, using statistical methods such as Poisson and negative binomial regression or generalized linear models to predict cycling activity based on infrastructure characteristics, land use, and sociodemographic factors \citep{griswold2011pilot, winters2016bike, lin2020modeling, livingston2021predicting}. 
While these methods provide baseline insights, they struggle with spatial and temporal sparsity present in the bicycling data, as they require substantial labeled data and fail to account for network scale spatial interdependencies. 
For instance, \citep{salon2016estimating} 
Salon demonstrated that, in networks with fewer than 10\% of monitored segments, models exhibited significant error rates due to their inability to propagate information across unobserved links. 
Machine learning techniques such as random forests and gradient boosting have improved prediction accuracy by incorporating non-linear relationships \citep{may2008vector, fu2016vehicle} but they too typically require sufficient labeled data across the network to generalize effectively and remain constrained by their inability to model spatial connectivity inherent to transportation networks.
To address spatial connectivity, Graph Neural Networks offer a promising approach by modeling network topology explicitly.

\subsection{Graph Neural Networks}
Graph Neural Networks (GNNs) have emerged as powerful tools for modeling transportation networks, given their capacity to exploit the underlying graph structure of road segments and intersections (Anonymous, 2024). 
In motorised traffic domains, GNN variants such as Diffusion Convolutional Recurrent Neural Networks (DCRNN) \citep{li2017diffusion} and Spatio-Temporal Graph Convolutional Networks (STGCN) \citep{yu2017spatio} have  advanced traffic flow and speed prediction by capturing spatial and temporal dependencies. 
Attention-based GNN models such as Graph Attention Networks (GAT) and their adaptations \citep{zhang2018gaan} have further improved performance by dynamically weighting important edges in the network.
Recent applications have begun to explore GNNs for bicycle and pedestrian traffic \citep{rahmani2023graph, miah2023estimation}, yet widespread adoption is hindered by the lack of sufficiently dense sensor data. 
Our prior work has shown that Graph Convolutional Networks (GCNs) can effectively model link-level bicycling volumes up to 80\% data sparsity (Anonymous, 2024). However, GNNs face significant challenges in bicycling networks with sparse sensor coverage as the limited labeled data hinder their generalization. 
Therefore, an optimized sensor placement is required to provide sufficient labeled data for enhanced data and effective prediction \citep{bhowmick2023systematic}.

\subsection{The Need for Optimized Sensor Placement}
Acknowledging the currently very limited coverage of bicycle count locations and amid growing interest in increasing the number of count locations in cities globally \citep{miah2022challenges, bhowmick2023systematic}, developing an optimized sensor placement strategy for increasing the accuracy of link-level bicycling volume estimation is a critical challenge.  
While cities are increasingly investing in bicycle counters to support urban planning and promote sustainable mobility, the absence of optimized sensor placement methods limits the effectiveness of data collection, particularly in sparse networks like Melbourne’s, where only 141 out of 15,933 segments are monitored (resulting in 99\% sparsity) \citep{schoner2021prioritizing, eurovelo2024cycling}. 
This gap poses a critical challenge for accurate link-level bicycling volume estimation, as capturing network-wide spatial dependencies requires sufficient labeled data which is often unavailable due to the limited number of sensors and their uneven distribution across road types (Section~\ref{sec: implication}).

In motor transportation, sensor placement has a well-established history. 
They have evolved from manual selection by traffic engineers in the 1990s to more systematic approaches by the 2000s \citep{guide2001traffic, meyer2016transportation}. 
Early methods relied on heuristic techniques - centrality measures such as betweenness centrality to identify critical locations for monitoring traffic flow \citep{zhao2017network}, followed by optimization techniques such as greedy algorithms to maximize coverage \citep{fakhouri2020multi}. 
Over time, these methods advanced with the integration of data-driven approaches, such as reinforcement learning (RL) for optimizing sensor locations in traffic flow prediction \citep{wang2019reinforcement} and graph neural networks (GNNs) for tasks such as traffic signal control by modeling network topology \citep{chen2020toward}, with RL also enhancing traffic light control through fuzzy inference systems \citep{kumar2020fuzzy}. 
Today, motor transportation benefits from extensive sensor coverage in many cities, enabling robust data collection and the application of advanced methods to optimize placement \citep{elassy2024intelligent}. 

In contrast, sensor placement in bicycle transportation has evolved more slowly due to the historical focus on motor traffic. 
Early approaches were ad hoc, placing counters at high-traffic areas such as popular bike paths without systematic analysis which lead to significant biases in data coverage \citep{pucher2008cycling}. 
More recently, cities have adopted policy-driven placement, installing counters on new bike lanes to monitor usage \citep{eurovelo2024cycling}, yet this approach does not address the challenge of network-wide estimation in sparse settings. 
Currently, bicycle networks remain severely under-monitored, with uneven sensor distribution. 
For example as shown in Section~\ref{sec: implication}, Melbourne’s network has 67 sensors on protected bike lanes but only 10 on arterial roads and 1 on local roads. 
While RL and GNNs have been applied in motor transportation, no prior studies have combined these approaches to optimize sensor placement for bicycling volume estimation, making this an underexplored area for research.

To date, no systematic approach has been developed to optimize sensor placement for link-level bicycle volume estimation, leaving a critical gap that limits the predictive performance of models like GNNs in sparse settings. 
Our study addresses this gap by introducing the INSPIRE-GNN framework which leverages RL and GNNs to strategically place sensors, ensuring effective data collection for enhanced volume prediction in sparse bicycling networks.


\subsection{Reinforcement Learning}
Reinforcement Learning (RL) has shown great potential in addressing optimization problems within transportation systems due to its ability to learn optimal policies through interaction with the environment. 
Applications of RL include traffic signal control \citep{gao2017adaptive}, dynamic routing \citep{pan2023deep} and resource allocation \citep{haydari2020deep}. 
More general frameworks for graph-based RL have emerged to handle a wide array of combinatorial optimization tasks including route planning and network design. 
A recent comprehensive survey \citep{darvariu2024graph} unifies these perspectives, illustrating how RL can leverage Graph Neural Networks (GNNs) to model topological and contextual information and then learn policies to solve complex discrete optimization problems over graphs.  Recent studies have started to integrate Graph Neural Networks with Reinforcement Learning to tackle complex problems that involve both spatial dependencies and sequential decision-making. 
For example, \citet{chen2020toward} proposed a GNN-RL framework for adaptive traffic signal control, capturing the network topology and learning optimal control policies. 
Similarly, \citet{wang2024large} employed GNNs within an RL framework for traffic flow optimization, demonstrating improved performance over traditional methods. 
Despite these advances, RL has not been widely applied to sensor placement in bicycling networks, where sparse sensor coverage necessitates strategic data collection to provide the labeled data needed for GNN-based volume predictions.

\paragraph{\textbf{Our Contribution:}} To the best of our knowledge, no prior work has combined GNNs with RL to optimize sensor placement in bicycling networks with sparse sensor coverage. 
Our work addresses this critical gap by introducing \networkName{}, an RL-enhanced GNN framework that integrates Graph Convolutional Networks (GCNs) and Graph Attention Networks (GATs) with a Deep Q-Network (DQN)-based RL agent to strategically place sensors, thereby improving link-level volume estimation.
This integration leverages GNNs' ability to model complex spatial relationships and the capability of RL  to learn optimal policies for sequential decision-making.


\section{Methodology} \label{sec:methodology}

In this section, we present the detailed methodology underlying \networkName, our novel framework for optimizing sensor placement in urban bicycling networks with sparse sensor coverage, using City of Melbourne’s bicycling network as our case study. With only 141 sensors across 15,933 segments, Melbourne exemplifies extreme sparsity (approximately 99\% of unmonitored segments) which severely limits accurate bicycle volume estimation. 
Building on prior work demonstrating GNNs’ limitations under such sparsity (Anonymous, 2024), the \networkName{} framework introduces RL to strategically select sensor locations, addressing gaps in traditional and optimization-based methods. 
Our approach combines a hybrid Graph Neural Network (GNN) model with a Reinforcement Learning (RL) agent to address the sparse sensor coverage challenge and enhance bicycle volume estimation accuracy

\subsection{Problem Formulation}

Accurately estimating bicycling volumes on each link (road or path segment) within an urban network is essential for informed transport planning and infrastructure development. 
However, the vast majority of links lack direct bicycling count data, likely due to a combination of factors, including financial constraints and the limited funding for active transport infrastructure, which can restrict the installation of widespread sensor networks and therefore, resulting in high data sparsity. 
To address this challenge, we aim to strategically place additional sensors to maximize the improvement in link-level bicycling volume estimation across the entire network.

We represent the urban bicycling network as an undirected graph \( \mathbf{G} = (\mathbf{V}, \mathbf{E}) \), where \( V = \{1, 2, \dots, N\} \) is the set of nodes, each corresponding to a road/path segment and \( E \subseteq V \times V \) is the set of edges, representing the connectivity between segments (i.e., if two segments are directly connected). 
Each node \( i \in V \) is associated with a feature vector \( \mathbf{x}_i \in \mathbb{R}^d \), capturing attributes - road type, slope, infrastructure characteristics, and a true bicycling count value \( y_i \in \mathbb{R} \), known only for nodes with sensors.

Let \( \mathcal{S}_{\text{existing}} \subset V \) denote the set of nodes with existing sensor, and \( \mathcal{S}_{\text{unlabeled}} = V \setminus \mathcal{S}_{\text{existing}} \) denote the nodes without sensors. 
In the City of Melbourne's bicycling network, \( |\mathcal{S}_{\text{existing}}| = 141 \) and \( |V| = 15,933 \), resulting in approximately 99\% data sparsity. 
Our primary objective is to select a subset \( \mathcal{S}_{\text{new}} \subseteq \mathcal{S}_{\text{unlabeled}} \) of \( K \) nodes to place additional sensors, minimizing the overall prediction error of bicycling volume estimates across all unlabeled nodes nodes. Formally, we aim to solve:
\begin{equation} \label{optimation_equation}
    \underset{\mathcal{S}_{\text{new}} \subseteq \mathcal{S}_{\text{unlabeled}}, |\mathcal{S}_{\text{new}}| = K}{\text{min}} \, \mathcal{F} \left( \hat{Y} \, \middle| \, S_{\text{train}} = \mathcal{S}_{\text{existing}} \cup \mathcal{S}_{\text{new}} \right)
\end{equation}

where, \( \hat{Y} = \{ \hat{y}_i : i \in \mathcal{S}_{\text{unlabeled}} \} \) is the set of estimated bicycling volumes for unlabeled nodes, \( \mathcal{F}(\cdot) \) is the evaluation metric (Mean Squared Error) and \( S_{\text{train}} \) is the set of nodes used for training the estimation model. 
Thus, our framework strategically selects sensor placements that minimize prediction error and improve the accuracy of bicycling volume estimation across the entire network.

\begin{figure*}[t]
    \centering
    \includegraphics[width=.8\linewidth]{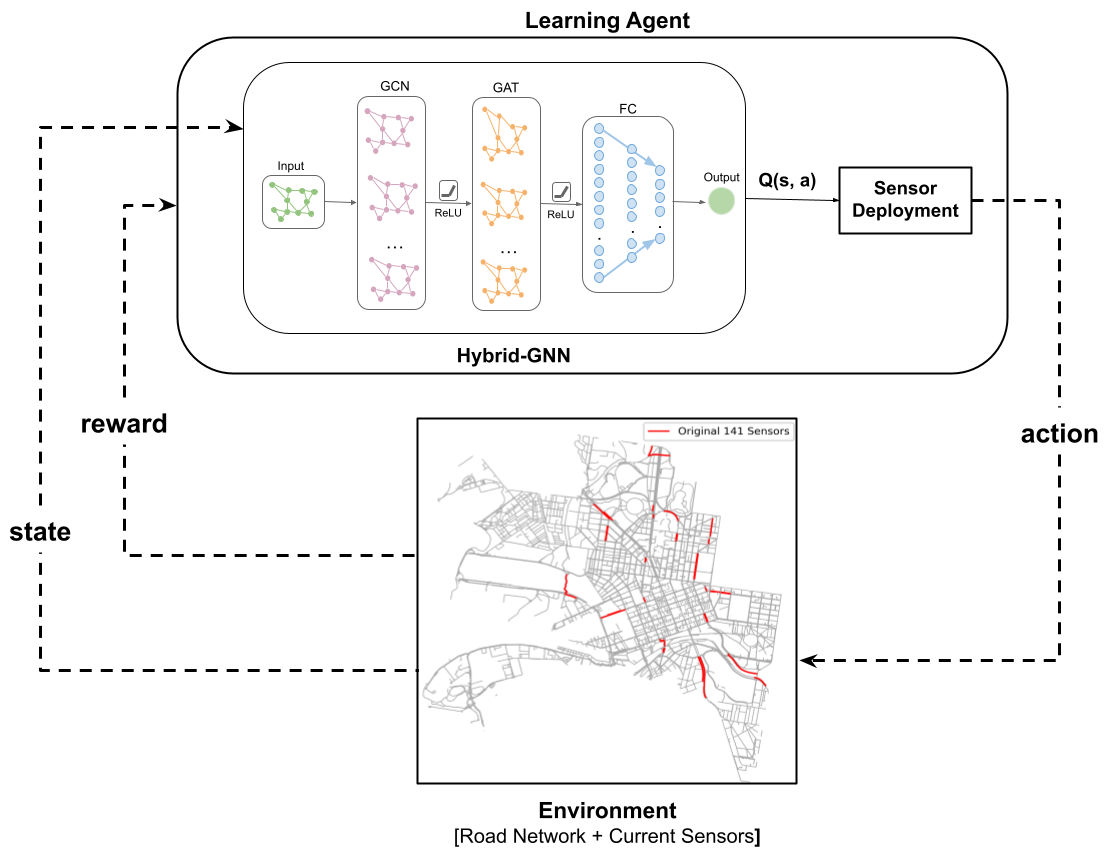}
    
    \caption{Reinforcement learning–based sensor placement workflow in INSPIRE‑GNN. The current bicycling network graph with existing sensors constitutes the \emph{state}, which is processed by the Hybrid‑GNN inside a DQN agent to estimate Q(s,a). The agent’s \emph{action}—selecting a new sensor location—is applied to the environment, and the reduction in prediction error ($\Delta\,\mathrm{MSE}$) obtained after retraining serves as the \emph{reward}, closing the feedback loop. This process iterates until the desired number of additional sensors is deployed.}
\label{fig:inspire-gnn}
\end{figure*}

\subsection{The \networkName{} Framework}
To address the optimization problem defined in \autoref{optimation_equation}, we propose the \textbf{\networkName} (\textbf{In}telligent \textbf{S}ensor \textbf{P}lacement to \textbf{I}mprove Sparse Bicycling Network Prediction via \textbf{Re}inforcement Learning Boosted Graph Neural Networks) framework.  
\networkName{} integrates two neural network architectures - a Graph Convolutional Network-Graph Attention hybrid Network (GCN-GAT) and a Deep Q-Network (DQN) based Reinforcement Learning (RL) agent as shown in Fig.~\ref{fig:inspire-gnn}.  
GCN-GAT is used for bicycling volume estimation, while the DQN adaptively learns optimal sensor placement strategies. 
DQN aims to maximize the GCN-GAT's performance by strategically selecting sensor locations.

Below, we detail the two main components of the \networkName{} framework - the \emph{Hybrid GNN Architecture} pipeline responsible for learning to predict link-level bicycling volumes, and the \emph{Reinforcement Learning (RL) Architecture} process, which focuses on optimally selecting sensor locations to improve overall prediction accuracy.

\subsubsection{Hybrid GNN (GCN+GAT) Architecture and Training} \label{sec:gnn_architecture_training}

The Hybrid GNN model in \networkName{} effectively learns representations from the bicycling network graph \( G = (V, E) \) and node features \( \mathbf{x}_i \) by combining Graph Convolutional Network (GCN) layers and Graph Attention Network (GAT) layers, leveraging their complementary strengths to capture both local connectivity and global structural patterns in the bicycling network. 
Our primary goal is to learn a function
\begin{equation} \label{hybrid-gnn}
    f : (X, E) \mapsto \hat{Y}
\end{equation}
that estimates the bicycling volume \(\hat{y}_i\) for each node \(i\).

\paragraph{Key Architectural Components: } 
\begin{itemize}
    
    \item  \textbf{GCN Layers with Residual Connections:} We stack GCN layers to capture localized structural information (i.e., how volumes propagate in immediate neighborhoods). 
    To stabilize training and enable deeper architectures, each GCN layer is equipped with residual connections \citep{he2016deep} and layer normalization \citep{ba2016layer}. This preserves information from earlier layers and mitigates vanishing gradients.
    
    \item \textbf{Edge Feature Encoding:} Since edges can hold critical data, we transform raw edge attributes into a learnable representation. 
    A trainable linear mapping (followed by non-linear activations) allows the model to incorporate domain-specific relationships at the edge level.

    \item \textbf{GAT Layers with Edge-Aware Attention:} After transforming the edge features, we integrate them into GAT layers \citep{velickovic2017graph}. 
    By attending to each edge’s importance, the network adaptively weighs the most influential connections, particularly valuable for identifying critical links in the cycling network.

    \item \textbf{Global Readout and Prediction:} The final node embeddings from the stacked GCN and GAT layers are aggregated via global mean pooling and global max pooling. 
    Their concatenation forms a comprehensive graph-level embedding, which is then fed into fully connected layers to produce final volume estimates, \(\hat{y}_i\).

\end{itemize}

\paragraph{Mathematical Formulation:} 
Let $\mathbf{X} \in \mathbb{R}^{N \times d}$ be the node features matrix and $\mathbf{E} \in \mathbb{R}^{|\mathcal{E}| \times d_e}$ the initial matrix of edge attributes. We denote the adjacency matrix by $\mathbf{A}$. The forward pass of our hybrid GNN proceeds as follows: 

\[
\mathbf{H}^{(0)} = \mathbf{X},
\]
\[
\mathbf{H}^{(1)} = \mathrm{ReLU}\Bigl( \mathrm{LayerNorm}\bigl( \mathrm{GCN}(\mathbf{H}^{(0)}, \mathbf{A}) \bigr) \Bigr) + \mathbf{H}^{(0)},
\]
\[
\mathbf{E}' = \mathrm{ReLU}\bigl( \mathbf{W}_e \, \mathbf{E}^{(0)} \bigr),
\]
\[
\mathbf{H}^{(2)} = \mathrm{ReLU}\Bigl( \mathrm{LayerNorm}\bigl( \mathrm{GAT}(\mathbf{H}^{(1)}, \mathbf{A}, \mathbf{E}') \bigr) \Bigr),
\]
\[
\mathbf{H}^{(3)}, \mathbf{A}', \mathbf{E}'' = \mathrm{TopKPooling}\Bigl( \mathbf{H}^{(2)}, \mathbf{A}, \mathbf{E}' \Bigr),
\]
\[
\mathbf{H}^{(L)} = \mathrm{GAT}\Bigl( \mathrm{GCN}\bigl( \mathbf{H}^{(3)} \bigr), \mathbf{A}', \mathbf{E}'' \Bigr),
\]
\[
\mathbf{H}_{\mathrm{global}} = \mathrm{GlobalMeanPool}\bigl( \mathbf{H}^{(L)} \bigr) \; \| \; \mathrm{GlobalMaxPool}\bigl( \mathbf{H}^{(L)} \bigr),
\]
\[
\hat{y}_i = f_{\mathrm{FC}}\bigl( \mathbf{H}_{\mathrm{global}} \bigr).
\]

We then train the model by minimizing the Mean Squared Error (MSE) between the predicted and true bicycling volumes:
\begin{equation}
\label{eq:gnn_loss}
L_{\mathrm{GNN}} = \frac{1}{|S_{\mathrm{train}}|} \sum_{i \in S_{\mathrm{train}}} \bigl( y_i - \hat{y}_i \bigr)^2.
\end{equation}

In the above formulation, we start by initializing the node features \(\mathbf{X}\) and edge features \(\mathbf{E}^{(0)}\), with \(\mathbf{A}\) representing the graph structure. 
The first GCN layer processes the node features and is equipped with a residual connection, ensuring that the initial information is preserved. 
Following this, the raw edge features are transformed via a learnable weight matrix \(\mathbf{W}_e\) and activated with ReLU to generate dynamic edge features \(\mathbf{E}'\). 
These transformed features are then incorporated into a GAT layer, which applies layer normalization and a ReLU activation to compute \(\mathbf{H}^{(2)}\). 
A pooling step - TopKPooling is employed to reduce the graph’s dimensionality, resulting in new representations \(\mathbf{H}^{(3)}\), \(\mathbf{A}'\), and \(\mathbf{E}''\). 
Further layers, potentially combining additional GCN and GAT blocks, yield the final node embeddings \(\mathbf{H}^{(L)}\). These embeddings are aggregated using both global mean and max pooling, concatenated to form \(\mathbf{H}_{\mathrm{global}}\), and finally passed through fully connected layers \(f_{\mathrm{FC}}\) to produce the predicted bicycling volumes \(\hat{y}_i\). 
The training objective minimizes the MSE over the training set \(S_{\mathrm{train}}\) comprising nodes with sensors, ensuring that the model learns an accurate mapping from the graph-structured inputs to the volume predictions.

\subsubsection{Reinforcement Learning – DQN Architecture and Training} \label{sec:rl_architecture_training}

Building on the Hybrid GNN’s capacity to estimate bicycling volumes, we treat sensor placement as a sequential decision-making problem. 
A Deep Q-Network (DQN) agent is trained to select new sensor locations that maximize the overall volume estimation performance improvement of the Hybrid GNN model across the network. 
At each decision step \(t\), the agent observes a \emph{state} \(s_t\) and selects an \emph{action} \(a_t\), receiving a corresponding \emph{reward} \(r_t\). 
This process continues until the predetermined number of sensor placements has been achieved.

\paragraph{State, Action, and Reward: }
\begin{itemize}
    \item \textbf{State Representation.} We define the state \(s_t\) to encapsulate the current configuration of sensors and the GNN’s representation of the network. In practice, we aggregate the GNN embeddings of the sensor-equipped nodes:
    \begin{equation}
    s_t = \mathrm{Agg}\Bigl( \{ \mathbf{h}_i : i \in S_{\mathrm{train}} \} \Bigr),
    \end{equation}
    where \(S_{\mathrm{train}} = S_{\mathrm{existing}} \cup S_{\mathrm{new}}\) and \(\mathrm{Agg}(\cdot)\) is an aggregation function (e.g., mean pooling) over the node embeddings \(\mathbf{h}_i\).

    \item \textbf{Action Space.} The action \(a_t\) corresponds to selecting one node from the unlabeled set:
    \begin{equation}
    \mathcal{A}_t = \{\, i \in V : i \notin S_{\mathrm{train}} \}.
    \end{equation}
    Each selected node \(a_t\) is “upgraded” to a sensor node.

    \item \textbf{Reward Function.} After selecting \(a_t\), the GNN is retrained (or fine-tuned) on the updated sensor set \(S_{\mathrm{train}}\). Let \(L_{\mathrm{val}}^{(t)}\) denote the GNN’s validation loss after the update. The reward is defined as:
    \begin{equation}\label{eq:dqn_reward}
    r_t = \Bigl( L_{\mathrm{val}}^{(t-1)} - L_{\mathrm{val}}^{(t)} \Bigr) + \beta \cdot r^{\mathrm{int}}_t,
    \end{equation}
    where \(r^{\mathrm{int}}_t\) is an intrinsic reward encouraging exploration, and \(\beta\) is a weighting factor.
\end{itemize}

\paragraph{DQN Formulation and Loss Function:}

We approximate the action-value function with a parametric function \( Q(s_t, a_t; \theta) \). The objective is to learn parameters \(\theta\) that maximize the expected cumulative reward:
\begin{equation}
\max_{\theta} \quad \mathbb{E}\Bigl[ \sum_{t=1}^{K} r_t \Bigr],
\end{equation}
where \(K\) is the total number of sensor placements. The DQN is trained by minimizing the Temporal Difference (TD) error. The TD target is defined as:
\begin{equation}
y_t = r_t + \gamma \max_{a' \in \mathcal{A}_{t+1}} Q\bigl( s_{t+1}, a'; \theta^- \bigr),
\end{equation}
with \(\gamma \in [0,1]\) as the discount factor and \(\theta^-\) representing the parameters of a periodically updated target network. The TD loss is then given by:
\begin{equation}\label{eq:dqn_loss}
L_{\mathrm{DQN}} = \mathbb{E}_{(s_t, a_t, r_t, s_{t+1})}\left[ \Bigl( y_t - Q(s_t, a_t; \theta) \Bigr)^2 \right].
\end{equation}

\paragraph{RL Search Approaches:}

To efficiently navigate the large action space, we compare three RL approaches:
\begin{enumerate}
    \item \textbf{Standard DQN.} The baseline approach without any advanced exploration enhancements.
    \item \textbf{Adaptive Epsilon-Greedy.} In this approach, the agent initially selects actions at random with probability \(\epsilon\), which decays over time to favor exploitation \citep{tokic2010adaptive}.
    \item \textbf{Curiosity-Driven Exploration.} Here, an intrinsic reward component \(r^{\mathrm{int}}_t\) is provided, which scales inversely with the frequency of encountering a given state \(s_t\), thereby encouraging exploration of novel regions.
\end{enumerate}

\paragraph{Training Procedure:}

The training of the DQN agent proceeds as follows:
\begin{itemize}
    \item \textbf{State Observation:} Compute the current state \(s_t\) using the aggregated embeddings of the sensor-equipped nodes.
    \item \textbf{Action Selection:} Select an action \(a_t \in \mathcal{A}_t\) using the \(\epsilon\)-greedy policy (incorporating the chosen exploration strategy).
    \item \textbf{Sensor Deployment and GNN Update:} Add the selected node \(a_t\) to \(S_{\mathrm{train}}\) and retrain (or fine-tune) the Hybrid GNN to incorporate the new sensor data, obtaining the updated validation loss \(L_{\mathrm{val}}^{(t)}\).
    \item \textbf{Reward Computation:} Calculate the reward \(r_t\) as in Equation~\eqref{eq:dqn_reward}.
    \item \textbf{Experience Replay:} Store the transition \((s_t, a_t, r_t, s_{t+1})\) in a replay buffer.
    \item \textbf{Parameter Update:} Sample a mini-batch of transitions from the replay buffer and perform a gradient descent step to minimize \(L_{\mathrm{DQN}}\) (Equation~\eqref{eq:dqn_loss}). Periodically update the target network parameters \(\theta^-\) with the current \(\theta\).
\end{itemize}

\paragraph{\textbf{Summary:}} By framing sensor placement as a DQN-based RL task, \networkName{} strategically selects sensor locations that yield maximum possible improvements in the Hybrid GNNs performance. 
The standard DQN baseline is enhanced by incorporating adaptive exploration strategies - Adaptive Epsilon-Greedy and Curiosity-Driven Exploration, to ensure that the agent effectively explores under-monitored regions of the network. 
This approach facilitates rapid improvements in volume estimation accuracy and provides a robust, data-driven method to mitigate the challenges of high data sparsity in urban bicycling networks.

\section{Experiments}
\label{sec:experiments}

In this section, we investigate the effectiveness of the proposed \networkName{} framework for enhancing link-level bicycling volume estimation under sparse sensor coverage. 
We compare different sensor placement strategies (both heuristic and RL-based) and evaluate the performance of our Hybrid GNN model against several baseline machine learning methods. 
We additionally conduct ablation studies to isolate the contributions of each framework component.

\subsection{Experimental Setup}
\label{sec:experimental_setup}

\begin{figure*}
    \centering
    \includegraphics[width=1\linewidth]{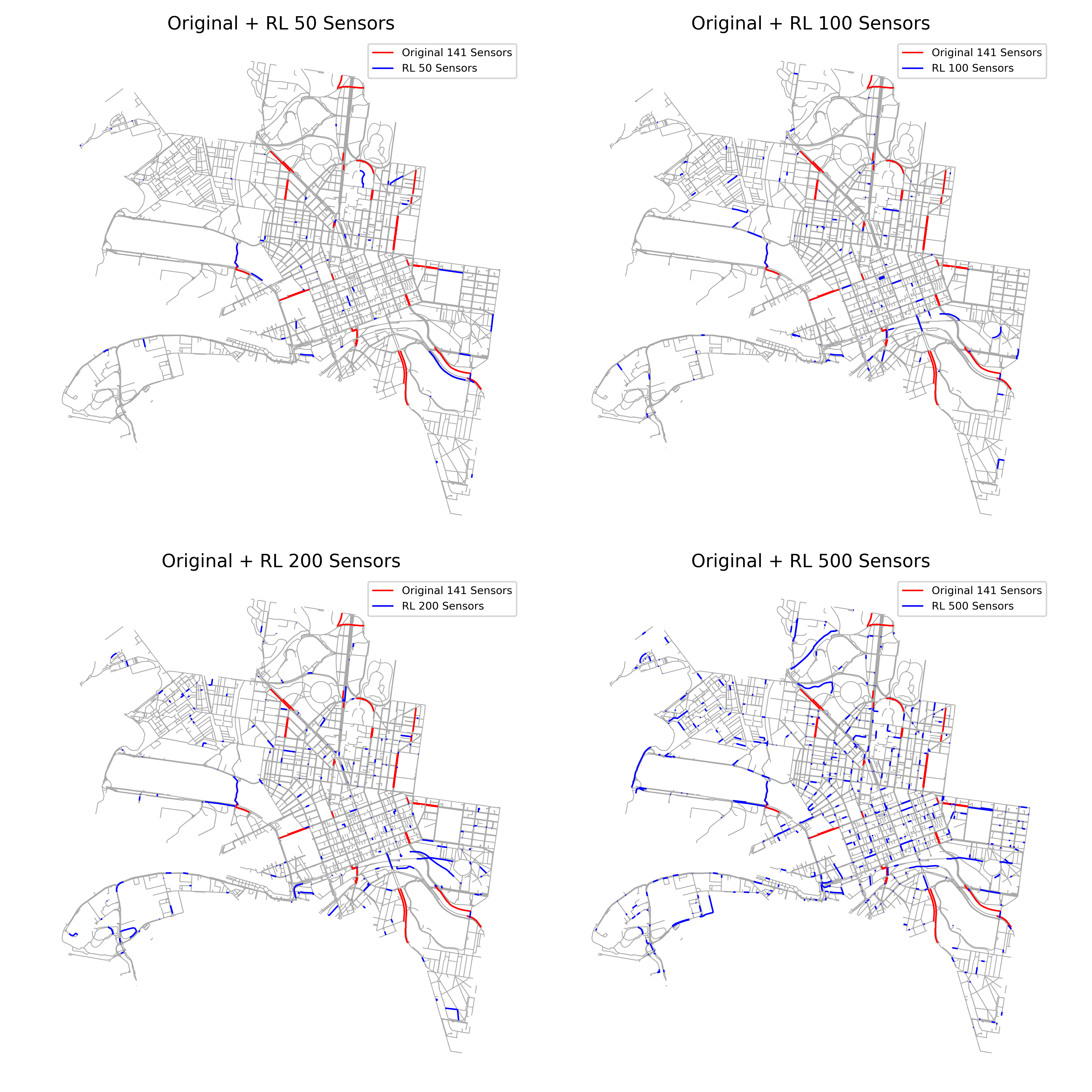}
    \caption{Spatial distribution of original and RL‑augmented sensor placements in Melbourne’s bicycling network. Each subplot overlays the existing 141 sensor locations (red) with additional sensors selected by the RL‑based Curiosity‑Driven Exploration policy (blue) for deployments of 50, 100, 200 and 500 sensors respectively.}
    \label{fig:melbourne_sensor_placement_subplots}
\end{figure*}

We focus our study on the bicycling network in the City of Melbourne, which features 15,933 road or path segments. 
Of these, 141 segments have bicycle count sensor coverage, strategically located along bike paths and bike lanes as shown in Fig. ~\ref{fig:melbourne road network vs original 141}. 
The primary goal of our framework is to identify the optimal subset of additional sensor locations from the remaining 15,792 unlabeled segments. 
This optimal subset is chosen such that it maximizes the link-level bicycle volume estimation performance by achieving the maximum possible reduction in prediction error. 
To evaluate \networkName{}, we simulate real-world conditions by deploying 50, 100, 200, and 500 new sensor locations as shown in Fig.~\ref{fig:melbourne_sensor_placement_subplots}, using Strava-derived bicycling volumes to gauge the effectiveness of different placement strategies.
To conduct the experiments, we used a combination of OpenStreetMap (OSM) data for network topology and Strava Metro data for bicycling volumes. 

\textbf{OpenStreetMap (OSM) Network Data:} We utilize OpenStreetMap (OSM) data to obtain the topology and infrastructure details of City of Melbourne’s bicycling network. 
Each of the 15,933 road/path segments is treated as a node in our graph, with edges denoting physical adjacency. 
Each node is enriched with attributes such as road type, slope, speed limits and bicycle infrastructure (e.g., bike lanes, shared paths, no bicycling infrastructure), enabling the model to account for diverse road characteristics. 
This data forms the foundation for capturing the spatial relationships between road or path segments.

\textbf{Strava Metro Data:}
Our study leverages aggregated and de-identified data from Strava Metro \citep{StravaMetro}, a widely used fitness tracking platform that has amassed an extensive repository of cycling activity data. Strava Metro data provides daily bicycling volumes for all 15,933 segments within the bicycle network in Melbourne. 
We derived the Annual Average Daily Bicycle (AADB) counts from these daily volumes, representing the average number of cyclists passing through each segment per day over a year. 

Strava data inherently have biases, it tends to overrepresent certain cyclist demographics, including recreational riders and fitness enthusiasts, while underrepresenting commuting cyclists and bicycling activity on quieter or less-traveled routes \citep{fischer2020comparing, lee2021strava}. 
Despite these limitations, Strava Metro’s data are often utilized due to their extensive geographic coverage (having data across the entire bikeable road network) and high spatial granularity, especially when compared against Melbourne’s extremely limited existing bicycling count sensor network comprising only 141 locations. 
Thus, Strava Metro data provide a valuable proxy for network-wide cycling volumes and are widely recognized as useful for modeling purposes in prior studies \citep{ferster2021mapping, nelson2023bicycle}. 

In our framework, the extensive spatial coverage of Strava Metro data allows us to simulate various realistic levels of data sparsity. 
This facilitates rigorous evaluation and benchmarking of the performance improvements provided by our RL-based sensor placement strategy, specifically designed to enhance link-level bicycling volume estimation in sparse sensor networks coverage.

\subsection{Dataset Splits}
To rigorously train and evaluate our predictive models, we split the dataset into three mutually exclusive subsets:

\begin{itemize}
    \item \textbf{Training Set (70\%):} Includes all segments currently equipped with physical sensors ($S_{\text{existing}}$) as well as any newly selected sensor locations ($S_{\text{new}}$) identified by the RL agent. The Hybrid GNN model is trained on this set.
    \item \textbf{Validation Set (15\%):} Consists of 15\% of the unlabeled segments ($S_{\text{unlabeled}}$). This set guides hyperparameter tuning and helps monitor for potential overfitting during training.
    \item \textbf{Test Set (15\%):} The remaining 15\% of unlabeled segments are reserved for final performance assessment. Neither model nor hyperparameters are adjusted based on test outcomes to ensure an unbiased evaluation of generalization.
\end{itemize}
All three subsets are strictly non-overlapping, ensuring that no segment appears in more than one set. 
To further enhance robustness, each experiment is repeated over 5 independent runs with distinct random seeds. 
We report the mean performance across these runs to mitigate the effects of stochastic variability in the training process, thereby yielding a more reliable performance estimate.

\subsection{Baseline Selection Policy} \label{sec:baseline_selection}

To assess the effectiveness of \networkName{} in determining optimal sensor placements, we compare its learning-based policy against several rule-based selection strategies. While these baselines do not adaptively learn from observed data, they offer common, well-understood heuristics for sensor placement:
\begin{itemize}
    \item \textbf{Random Selection:} Chooses road segments at random from \(S_{\mathrm{unlabeled}}\). This serves as a simple, non-informed baseline.
    \item \textbf{Betweenness Centrality:} Places sensors on nodes with the highest betweenness centrality, assuming nodes lying on many shortest paths are key to monitoring major cycling corridors.
    \item \textbf{Closeness Centrality:} Selects nodes with the highest closeness centrality, favoring centrally located segments that minimize average distance to all other network points.
    \item \textbf{Observed Bicycling Activity:} Allocates sensors to segments with the highest Strava-reported activity, prioritizing already busy links.
\end{itemize}
By comparing these traditional, rule-based approaches with \networkName{}, we can evaluate how much a data-driven and adaptive method improves sensor placement for accurate bicycle volume estimation.

\subsection{RL Approaches}
\label{sec:rl_approaches}

We implement three variants of Reinforcement Learning (RL) to identify how \emph{exploration strategies} influence sensor placement decisions:

\begin{enumerate}
    \item \textbf{Standard DQN (RL):} A baseline Deep Q-Network that relies solely on extrinsic rewards defined by improvements in the Hybrid GNN’s validation performance. This provides a starting point to gauge the impact of more sophisticated exploration.
    \item \textbf{RL with Adaptive Epsilon-Greedy:} Enhances the baseline DQN by adjusting the exploration rate \(\epsilon\) over time. Early in training, a higher \(\epsilon\) value encourages the agent to explore a wide range of nodes, gradually shifting toward exploitation of more promising sensor placements as learning progresses.
    \item \textbf{RL with Curiosity-Driven Exploration:} In addition to extrinsic rewards, this approach incorporates an intrinsic reward \(r_{t}^{\mathrm{int}}\) that is inversely proportional to the frequency of visiting a particular state.
    This mechanism incentivizes the agent to explore less-frequented areas of the network, potentially discovering high-impact sensor locations that standard exploration may overlook. This helps avoid prematurely converging to sub-optimal placements.
\end{enumerate}
These three RL approaches allow us to quantify how advanced exploration mechanisms affect the final sensor deployment strategy and the subsequent accuracy of link-level volume estimation.

\subsection{Baseline ML Models}
\label{sec:baseline_ml_models}

To assess the performance of our \emph{Hybrid GNN} architecture, we benchmark it against a range of established machine learning and deep learning models. This comparison clarifies whether leveraging the graph structure and attention mechanisms provides tangible benefits over conventional methods:

\begin{itemize}
    \item \textbf{Linear Regression:} A standard linear model that assumes a purely linear relationship between the input features and bicycling volumes.
    \item \textbf{Decision Tree:} A non-linear model that recursively partitions the feature space, enabling non-linear relationships but potentially overfitting if uncontrolled.
    \item \textbf{Random Forest:} An ensemble method that combines multiple decision trees via bootstrap aggregation, often improving robustness, generalization and reduce overfitting. 
    \item \textbf{Gradient Boosting:} A sequential ensemble technique that builds models incrementally, focusing on the errors of previous iterations.
    \item \textbf{Multi-Layer Perceptron (MLP):} A feedforward neural network capable of capturing nonlinearities through dense layers.
    \item \textbf{Convolutional Neural Network (CNN):} A deep learning which is typically employed for images, it can process spatial patterns if the network structure is interpreted in a grid-like manner.
\end{itemize}


\begin{table*}[t]
\centering
\caption{Comparison of Hybrid-GNN performance metrics for link-level bicycling volume estimation - trained under different sensor expansion scenarios (50, 100, 200, and 500 additional sensors) using Standard DQN, Adaptive Epsilon-Greedy and Curiosity-Driven Exploration policies, against the baseline trained with only the original 141 counters.}
\label{tab:results_rl_comparison_table}
\renewcommand{\arraystretch}{1.2} 
\begin{tabular}{l|c|c|cccc}
\textbf{Selection Method} & \textbf{\begin{tabular}[c]{@{}c@{}}Number of \\ Additional Nodes\end{tabular}} & \multicolumn{1}{l|}{\textbf{\begin{tabular}[c]{@{}l@{}}Total Nodes in \\ GNN Training\end{tabular}}} & \textbf{MSE Loss} & \textbf{RMSE} & \textbf{MAPE (\%)} & \textbf{MAE} \\ \hline
Original Counters & 0 & 141 & 1976.30 & 44.46 & 18.49 & 33.21 \\ \hline
Standard RL & \multirow{3}{*}{50} & \multirow{3}{*}{141 + 50} & 1029.04 & 32.78 & 4.36 & 20.98 \\
Adaptive Epsilon-Greedy &  &  & 1045.92 & 32.34 & 4.51 & 21.24 \\
Curiosity-Driven Exploration &  &  & \textbf{1020.48} & \textbf{32.00} & \textbf{4.31} & \textbf{20.94} \\ \hline
Standard RL & \multirow{3}{*}{100} & \multirow{3}{*}{141 + 100} & 991.79 & 31.93 & 5.36 & 20.20 \\
Adaptive Epsilon-Greedy &  &  & 1000.54 & \textbf{31.62} & 5.20 & 20.12 \\
Curiosity-Driven Exploration &  &  & \textbf{990.26} & 31.80 & \textbf{5.10} & \textbf{20.03} \\ \hline
Standard RL & \multirow{3}{*}{200} & \multirow{3}{*}{141 + 200} & 909.48 & 31.16 & 4.05 & 18.86 \\
Adaptive Epsilon-Greedy &  &  & 910.34 & 30.15 & \textbf{4.00} & \textbf{18.84} \\
Curiosity-Driven Exploration &  &  & \textbf{905.67} & \textbf{30.08} & 4.04 & 18.89 \\ \hline
Standard RL & \multirow{3}{*}{500} & \multirow{3}{*}{141 + 500} & 865.90 & 29.23 & 3.54 & \textbf{17.46} \\
Adaptive Epsilon-Greedy &  &  & 870.92 & 29.50 & 3.61 & 17.53 \\
Curiosity-Driven Exploration &  &  & \textbf{865.32} & \textbf{29.22} & \textbf{3.52} & 17.48 \\ \hline
\end{tabular}
\end{table*}

\begin{figure*}[t]
    \centering
    \includegraphics[width=1\linewidth]{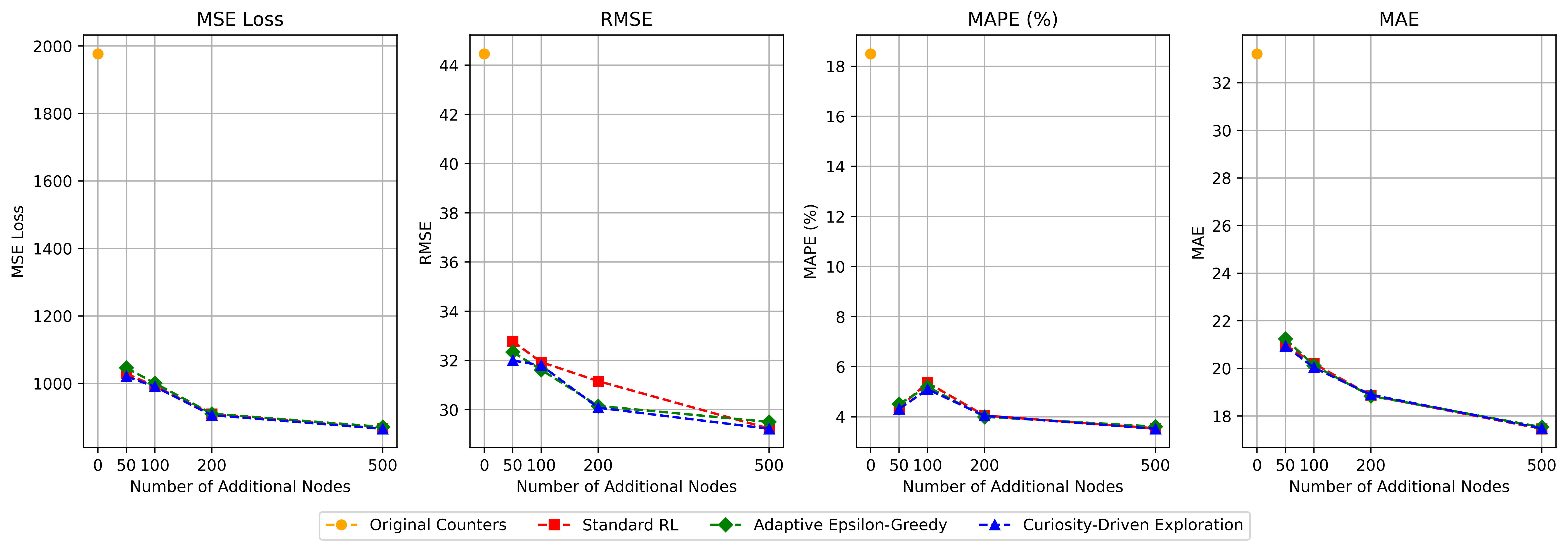}
    \caption{Trends in Hybrid-GNN performance metrics (MSE, RMSE, MAPE, and MAE) across different sensor expansion scenarios (50, 100, 200, and 500 additional sensors) using Standard DQN, Adaptive Epsilon-Greedy, and Curiosity-Driven Exploration policies, compared to the baseline model trained with only the original 141 counters.}
    \label{fig:results_rl_comparison_figure}
\end{figure*}


For each model, we evaluate performance on both the initial 141 sensor locations and under various sensor expansion scenarios (additional 50, 100, 200, and 500 sensors). This allows us to verify whether the Hybrid GNN’s specialized design, particularly its ability to exploit graph connectivity - consistently outperforms generic baselines when additional sensors become available.

\subsection{Performance Metrics}
\label{sec:performance_metrics}
We have used the following four commonly used metrics to evaluate the model performance:

\begin{itemize}
    \item \textbf{Mean Squared Error (MSE)}: penalizes large errors more heavily due to the squaring term.
    \[
    \mathrm{MSE} = \frac{1}{n} \sum_{i=1}^{n} (y_i - \hat{y}_i)^2
    \]

    \item \textbf{Root Mean Squared Error (RMSE)}: offers an interpretable error scale in the same units as the target variable.
    \[
    \mathrm{RMSE} = \sqrt{\mathrm{MSE}}
    \]

    \item \textbf{Mean Absolute Error (MAE)}: provides a straightforward interpretation of average deviation without heavily penalizing outliers.
    \[
    \mathrm{MAE} = \frac{1}{n} \sum_{i=1}^{n} \bigl|y_i - \hat{y}_i \bigr|
    \]
    
    \item \textbf{Mean Absolute Percentage Error (MAPE)}: gives a percentage-based measure of prediction error, useful for comparing model performance across segments with varying volume scales.
    \[
    \mathrm{MAPE} = \frac{100\%}{n}\sum_{i=1}^{n} \left| \frac{y_i - \hat{y}_i}{y_i} \right|
    \]
\end{itemize}

In our results, we report and compare all four complementary metrics to provide a comprehensive view of predictive performance and capture different aspects of estimation accuracy

\subsection{RL-Based Sensor Placement}
Traditional bicycling sensor placement strategies commonly focus on high-usage locations or newly constructed infrastructure, potentially neglecting critical yet underrepresented segments. 
To evaluate and understand how the RL-based approach strategically fills these coverage gaps, we systematically compared RL-selected sensor placements against the existing 141 sensor locations across Melbourne’s bicycling network. We classified and analyzed the placements using the following road-segment classifications, adapted from \citep{pearson2025gender}:

\begin{itemize}
    \item \textbf{Protected Bike Lanes:} Physically separated lanes dedicated exclusively to cyclists, providing enhanced safety from motorized traffic. Typically they are installed along major commuting corridors.
    \item \textbf{Painted Bike Lanes (Arterial and Collector Roads):} Bicycle lanes designated by painted markings without physical separation. Arterial roads generally accommodate high traffic volumes at higher speeds, while collector roads generally manage moderate traffic volumes connecting arterial roads with local streets.
    \item \textbf{Off-Road Bike Paths:} Paths completely separated from motorized traffic, commonly found through parks, greenways, frequently utilized by recreational cyclists.
    \item \textbf{Local Roads (Mixed Traffic or Sharrows):} Roads shared by bicycles and vehicles, either marked with shared-lane markings (sharrows) or without explicit bicycling infrastructure, typically serving residential and short-distance trips.
    \item \textbf{Arterial Roads (Mixed Traffic):} Busy, high-volume roads without dedicated cycling facilities, exposing cyclists directly to motorized traffic, typically less attractive due to perceived and actual safety risks.
    \item \textbf{Other (Including minor shared paths or informal routes):} Short connectors or informal cycling pathways not fitting clearly within the above classifications.
\end{itemize}

We quantitatively compared the distribution of sensors across these road categories before and after RL-driven sensor placements (50, 100, 200, and 500 additional sensors). This analysis highlights INSPIRE-GNN's strategic placement logic, targeting important yet underrepresented infrastructure segments to improve network representativeness and predictive accuracy, thereby complementing traditional sensor deployment practices.

\begin{table*}[t]
\centering
\caption{Comparison of baseline and RL-driven sensor placement strategies for additional deployments of 50, 100, 200, and 500 sensors. Performance is evaluated on link‑level bicycling volume estimation in Melbourne’s network, reporting MSE, RMSE, MAPE, and MAE.}

\label{tab:baseline_policy_comparison}
\renewcommand{\arraystretch}{1.2} 
\begin{tabular}{l|c|c|cccc}
\textbf{Selection Method} & \textbf{\begin{tabular}[c]{@{}c@{}}Number of \\ Additional Nodes\end{tabular}} & \textbf{\begin{tabular}[c]{@{}c@{}}Total Nodes in \\ GNN Training\end{tabular}} & \textbf{MSE Loss} & \textbf{RMSE} & \textbf{MAPE} & \textbf{MAE} \\ \hline
Betweenness Centrality & \multirow{5}{*}{50} & \multirow{5}{*}{141 + 50} & 1533.25 & 39.16 & 11.63 & 26.22 \\
Closeness Centrality &  &  & 2694.97 & 51.91 & 16.90 & 36.71 \\
Strava Activity &  &  & 5562.82 & 74.58 & 20.69 & 47.79 \\
Random Selection &  &  & 2217.36 & 47.09 & 14.58 & 32.08 \\
Curiosity-Driven Exploration &  &  & \textbf{1020.48} & \textbf{32.00} & \textbf{4.31} & \textbf{20.94} \\ \hline
Betweenness Centrality & \multirow{5}{*}{100} & \multirow{5}{*}{141 + 100} & 1806.87 & 42.51 & 15.04 & 32.52 \\
Closeness Centrality &  &  & 2200.15 & 46.91 & 15.02 & 33.00 \\
Strava Activity &  &  & 8023.02 & 89.57 & 26.65 & 60.77 \\
Random Selection &  &  & 1900.51 & 43.59 & 12.23 & 28.71 \\
Curiosity-Driven Exploration &  &  & \textbf{990.26} & \textbf{31.80} & \textbf{5.10} & \textbf{20.03} \\ \hline
Betweenness Centrality & \multirow{5}{*}{200} & \multirow{5}{*}{141 + 200} & 2079.14 & 45.60 & 13.64 & 30.11 \\
Closeness Centrality &  &  & 3000.51 & 54.78 & 18.91 & 38.68 \\
Strava Activity &  &  & 6896.09 & 83.04 & 28.82 & 61.98 \\
Random Selection &  &  & 1028.62 & 32.07 & 8.38 & 20.45 \\
Curiosity-Driven Exploration &  &  & \textbf{905.67} & \textbf{30.08} & \textbf{4.04} & \textbf{18.89} \\ \hline
Betweenness Centrality & \multirow{5}{*}{500} & \multirow{5}{*}{141 + 500} & 2167.15 & 46.55 & 14.53 & 31.97 \\
Closeness Centrality &  &  & 2368.84 & 48.67 & 13.84 & 31.07 \\
Strava Activity &  &  & 5409.96 & 73.55 & 28.66 & 57.82 \\
Random Selection &  &  & 1061.85 & 32.59 & 8.32 & 19.87 \\
Curiosity-Driven Exploration &  &  & \textbf{865.32} & \textbf{29.22} & \textbf{3.52} & \textbf{17.48} \\ \hline
\end{tabular}
\end{table*}


\section{Results and Discussion}
In this section, we present and discuss the results of extensive experiments evaluating the effectiveness of the proposed \networkName{} framework. 
\begin{itemize}
    \item We first assess the impact of deploying additional sensors selected through various reinforcement learning (RL) exploration strategies on the performance of our Hybrid-GNN model.
    \item Next, we compare our best performing RL-based sensor placement policy against traditional heuristic-based sensor selection policies.
    \item We then benchmark the predictive capability of our Hybrid-GNN against popular machine learning and deep learning models across different sensor placement scenarios.
    \item Additionally, an ablation study is conducted to quantify the contributions of each component within the \networkName{} framework.
    \item Finally, we discuss the practical implications of our RL-driven sensor placement approach for urban transportation planners, emphasizing its strategic benefits for improving bicycle infrastructure planning.
\end{itemize}

\subsection{Impact of Additional Sensor Placements and RL Exploration Strategies}
\label{sec:results_rl_comparison}

Table~\ref{tab:results_rl_comparison_table} and Figure~\ref{fig:results_rl_comparison_figure}  show the performance of our Hybrid-GNN model under various sensor deployment scenarios, comparing three Reinforcement Learning (RL) approaches — \emph{Standard RL}, \emph{RL with Adaptive Epsilon-Greedy}, and \emph{RL with Curiosity-Driven Exploration} — against a baseline scenario that uses only the \emph{original 141 counters}. 
Each row corresponds to a different number of newly placed sensors (50, 100, 200, or 500) added to the training set.

\textit{Performance Gains Over the Original Counters:} All three RL methods substantially outperform the baseline (with only 141 counters).
For instance, when adding 50 sensors, the MSE Loss decreases from 1976.30 to as low as 1020.48, while RMSE also drops from 44.46 to 32. 
Similar trends hold for MAE and MAPE, reflecting consistent improvements across all error metrics. 
These findings confirm that even a modest sensor expansion (e.g., 50 new sensors) can notably enhance the Hybrid-GNN’s predictive performance compared to relying solely on the sparse, existing counters.

\textit{Comparison of RL Exploration Strategies:} Among the three RL approaches, Curiosity-Driven Exploration generally provides the best performance across most metrics and sensor deployment sizes. 
For example, with 50 additional sensors, curiosity-based exploration yields an MSE of 1020.48, outperforming both Standard RL (1029.04) and Adaptive Epsilon-Greedy (1045.92). 
This pattern continues as the number of sensors grows to 500, where Curiosity-Driven Exploration slightly edges out the other RL variants on MSE, RMSE, and MAPE. 
These results suggest that the intrinsic reward mechanism successfully encourages the RL agent to explore less-monitored regions of the network, discovering sensor placements that maximize overall predictive accuracy.

Overall, these results validate the effectiveness of the Hybrid-GNN model when augmented with additional sensors selected via an RL-based approach. 
Among the three RL strategies, Curiosity-Driven Exploration offers the best performance across most scenarios. 
Strategic sensor expansions using Curiosity-Driven Exploration lead to significant predictive improvements, particularly evident even at moderate sensor expansions (e.g., 50-100 sensors) and will therefore serve as the primary sensor placement policy for subsequent comparisons with heuristic-based selection methods and machine learning baselines. 
These promising results motivate further comparisons with traditional heuristic methods to confirm the superiority of our RL-based sensor placement strategy.

\begin{table*}[t]
\centering
\caption{Comparison of Hybrid GNN vs. Baseline ML/DL Models Under Different Sensor Expansion Scenarios}
\label{tab:baseline_model_comparison}
\renewcommand{\arraystretch}{1.2} 
\begin{tabular}{l|c|c|cccc}
\textbf{Model} & \textbf{\begin{tabular}[c]{@{}c@{}}Number of \\ Additional Nodes\end{tabular}} & \textbf{\begin{tabular}[c]{@{}c@{}}Total Nodes for \\ Model Training\end{tabular}} & \textbf{MSE Loss} & \textbf{RMSE} & \textbf{MAPE (\%)} & \textbf{MAE} \\ \hline
Decision Tree & \multirow{6}{*}{0} & \multirow{6}{*}{141} & 4766.1 & 69.04 & 52.39 & 1568.7 \\
Random Forest &  &  & 4671.4 & 68.35 & 54.56 & 1601.67 \\
Gradient Boosting &  &  & 3551.55 & 59.59 & 47.29 & 1213.34 \\
MLP &  &  & 4128.71 & 64.26 & 51.1 & 1353 \\
CNN &  &  & 3530.36 & 59.42 & 57.27 & 1962.82 \\
GAT-GCN Hybrid &  &  & \textbf{1976.3} & \textbf{44.46} & \textbf{18.49} & \textbf{33.21} \\ \hline
Decision Tree & \multirow{6}{*}{50} & \multirow{6}{*}{141 + 50} & 2051.02 & 45.29 & 19.53 & 476.51 \\
Random Forest &  &  & 1989.88 & 44.61 & 22.49 & 708.42 \\
Gradient Boosting &  &  & 1912.41 & 43.73 & 22.67 & 749.67 \\
MLP &  &  & 2013.55 & 44.87 & 23.4 & 622.05 \\
CNN &  &  & 2175.41 & 46.64 & 43.37 & 439.94 \\
GAT-GCN Hybrid &  &  & \textbf{1020.48} & \textbf{32} & \textbf{4.31} & \textbf{20.94} \\ \hline
Decision Tree & \multirow{6}{*}{100} & \multirow{6}{*}{141 + 100} & 1431.49 & 37.83 & 15.21 & 329.7 \\
Random Forest &  &  & 1071.61 & 32.74 & 15.47 & 414.91 \\
Gradient Boosting &  &  & 1116.73 & 33.42 & 17.18 & 518.72 \\
MLP &  &  & 1441.38 & 37.97 & 19.3 & 522.12 \\
CNN &  &  & 1754.68 & 41.89 & 37.88 & 433.27 \\
GAT-GCN Hybrid &  &  & \textbf{990.26} & \textbf{31.8} & \textbf{5.1} & \textbf{20.03} \\ \hline
Decision Tree & \multirow{6}{*}{200} & \multirow{6}{*}{141 + 200} & 1408.58 & 37.53 & 15.48 & 343.79 \\
Random Forest &  &  & 1124.41 & 33.53 & 15.22 & 387.87 \\
Gradient Boosting &  &  & 1031.21 & 32.11 & 15.4 & 426.86 \\
MLP &  &  & 1143.95 & 33.82 & 16.24 & 374.43 \\
CNN &  &  & 1101.63 & 33.19 & 18.29 & 304.34 \\
GAT-GCN Hybrid &  &  & \textbf{905.67} & \textbf{30.08} & \textbf{4.04} & \textbf{18.89} \\ \hline
Decision Tree & \multirow{6}{*}{500} & \multirow{6}{*}{141 + 500} & 1599.06 & 38.85 & 15.03 & 350.02 \\
Random Forest &  &  & 1100.23 & 36.16 & 14.59 & 310.84 \\
Gradient Boosting &  &  & 999.38 & 33.99 & 14.66 & 374.87 \\
MLP &  &  & 1139.98 & 38.66 & 15.37 & 286.63 \\
CNN &  &  & 1057.6 & 35.28 & 17.15 & 272.83 \\
GAT-GCN Hybrid &  &  & \textbf{865.32} & \textbf{29.22} & \textbf{3.52} & \textbf{17.48} \\ \hline
\end{tabular}
\end{table*}

\begin{figure*}[!t]
    \centering
    \includegraphics[width=1\linewidth]{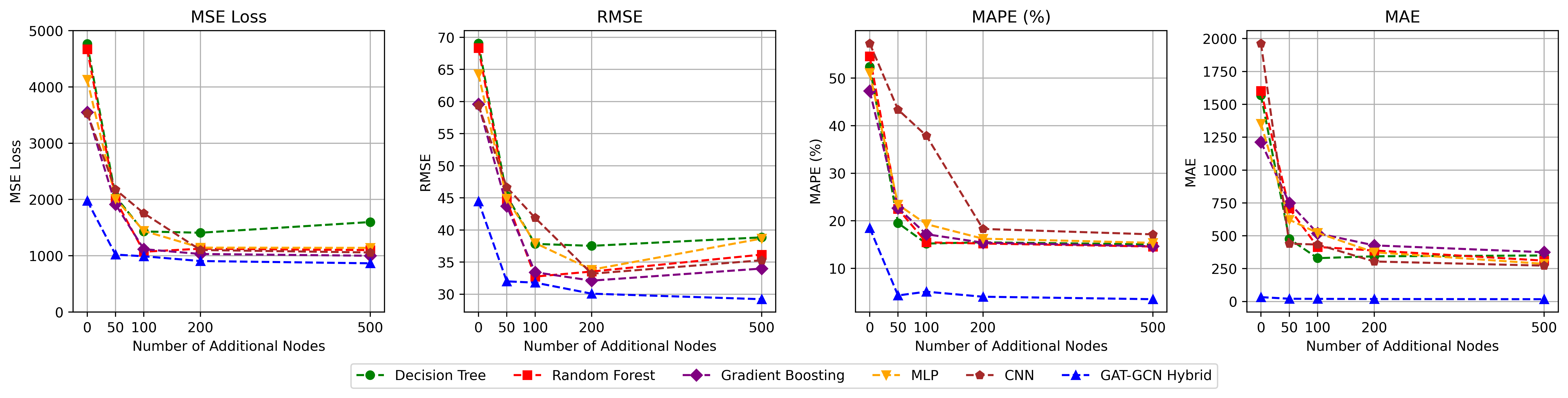}
    \caption{Comparison of the Hybrid GNN model against baseline machine learning and deep learning models for various sensor expansions (50, 100, 200, and 500 additional nodes), illustrating the Hybrid GNN’s consistently superior accuracy.}
    \label{fig:baseline_model_comparison_figure}
\end{figure*}


\subsection{Comparison of RL-Based Sensor Placement with Baseline Policies}
\label{sec:results_baseline_comparison}

To evaluate the effectiveness of our RL-based sensor placement strategy, we compare its performance with several traditional  heuristic selection policies. 
Specifically, we assess how the Hybrid-GNN model performs when trained with additional sensors - added using data-driven Curiosity-Driven Exploration policy (superior performance among the RL approaches as shown in 
Table~\ref{tab:results_rl_comparison_table}) in comparison to rule-based baseline policies. The baseline policies include Betweenness Centrality, Closeness Centrality, Strava Activity, and Random Selection. 

\begin{table*}[t]
\centering
\caption{Ablation Study Results for INSPIRE-GNN}
\label{tab:ablation_study_table}
\renewcommand{\arraystretch}{1.2} 
\begin{tabular}{l|l|cccc}
\textbf{Model Variant} & \textbf{Components Included} & \textbf{MSE Loss} & \textbf{RMSE} & \textbf{MAPE (\%)} & \textbf{MAE} \\ \hline
INSPIRE-GNN & \begin{tabular}[c]{@{}l@{}}Hybrid GAT-GCN + RL \\ with Curiosity-Driven Exploration\end{tabular} & \textbf{1020.48} & \textbf{32} & \textbf{4.31} & \textbf{20.94} \\ \hline
Without RL & \begin{tabular}[c]{@{}l@{}}Hybrid GAT-GCN + \\ Random Sensor Placement\end{tabular} & 2217.36 & 47.09 & 14.58 & 32.08 \\ \hline
Using Only GCN & \begin{tabular}[c]{@{}l@{}}GCN + RL \\ with Curiosity-Driven Exploration\end{tabular} & 1610.25 & 39.73 & 10.5 & 25 \\ \hline
Using Only GAT & \begin{tabular}[c]{@{}l@{}}GAT + RL \\ with Curiosity-Driven Exploration\end{tabular} & 1483.51 & 37.45 & 9.8 & 23.5 \\ \hline
\end{tabular}
\end{table*}

\begin{figure*}[t]
    \centering
    \includegraphics[width=1\linewidth]{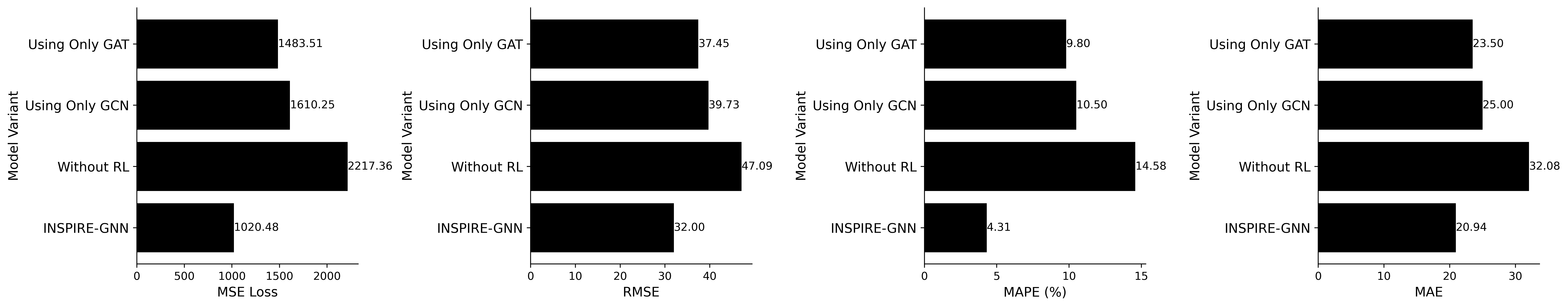}
    \caption{Performance Comparison of INSPIRE-GNN Variants on a sensor network comprising 141 original sensors and 50 additional sensors selected by the RL module. 
    The figure illustrates the impact of removing or isolating key components—such as the RL module, GCN layers, and GAT layers—on the model’s performance (MSE, RMSE, MAPE, MAE) in link-level bicycling volume estimation.}
    \label{fig:ablation_study_figure}
\end{figure*}

Table~\ref{tab:baseline_policy_comparison} presents the performance metrics — MSE, RMSE, MAPE and MAE — of the Hybrid-GNN model for sensor expansions of 50, 100, 200, and 500 additional nodes (resulting in training sets of 141 plus additional sensors). 
Across all deployment scenarios, the RL-based strategy (Curiosity-Driven Exploration) consistently outperforms the baseline policies. 
For example, with 50 additional sensors, the RL approach achieves an MSE of 1020.48, compared to 1533.25 for Betweenness Centrality, 2694.97 for Closeness Centrality, 5562.82 for Strava Activity, and 2217.36 for Random Selection. 
This pattern of improved performance is maintained across sensor expansions, with the RL approach yielding the lowest errors in all metrics at each sensor count.

These results demonstrate that the data-driven, adaptive RL strategy is more effective at identifying sensor placements that enhance the accuracy of link-level bicycling volume estimation than static, heuristic-based methods. 
The superior performance of the RL approach can be attributed to its ability to learn and exploit complex spatial relationships in the network, thereby selecting sensor locations that provide the most informative data for the Hybrid-GNN model.

\subsection{Comparison of Hybrid GNN with Other Baseline Models}
\label{sec:results_baseline_models}

Next, we benchmark the performance of our Hybrid GNN (GCN-GAT) model against multiple widely-used machine learning (ML) and deep learning (DL) approaches, including Decision Trees, Random Forests, Gradient Boosting, Multi-Layer Perceptrons (MLPs), and Convolutional Neural Networks (CNNs). We compare these methods under two primary scenarios:

\begin{itemize}
    \item The baseline scenario with only the original 141 sensor-equipped nodes.
    \item Sensor expansion scenarios involving the addition of 50, 100, 200, and 500 nodes, strategically selected using the Curiosity-Driven Exploration RL policy.
\end{itemize}

Table~\ref{tab:baseline_model_comparison} presents a detailed comparison of model performance across Mean Squared Error (MSE), Root Mean Squared Error (RMSE), Mean Absolute Percentage Error (MAPE), and Mean Absolute Error (MAE). Additionally, Fig.~\ref{fig:baseline_model_comparison_figure} visually illustrates the comparative trends in performance metrics across different sensor deployment levels.

The results clearly highlight the superior predictive performance of the Hybrid GNN model across all experimental scenarios and evaluation metrics. 
With the original 141 sensors, the Hybrid GNN achieves an MSE of 1976.30, considerably outperforming the next-best method, the CNN (3530.36 MSE) and Gradient Boosting (3551.55). 
Notably, traditional ML approaches such as Decision Trees, Random Forests, and MLPs exhibit substantially higher errors due to their limited ability to model the inherent spatial relationships of bicycling network data.

With incremental additions of strategically placed sensors via the Curiosity-Driven Exploration policy, the Hybrid GNN maintains its performance advantage. 
For instance, with an additional 50 sensors, the Hybrid GNN reduces its MSE to 1020.48, significantly outperforming Gradient Boosting (2051.02), Random Forest (1989.88), and CNN (2217.36). 
This advantage persists even when sensor placements increase to 500, with the Hybrid GNN attaining an MSE of 865.32, notably lower than Gradient Boosting (999.38), Random Forest (1100.23), and CNN (1057.6).

This consistent performance advantage underscores the Hybrid GNN’s effectiveness at capturing complex spatial patterns and dependencies in the bicycling network. 
Unlike traditional ML methods, which typically assume independence between data points, the Hybrid GNN explicitly leverages the connectivity and topology of the underlying bicycling network. 
The integration of graph convolutional layers with attention mechanisms allows it to dynamically prioritize important connections and relationships, which proves especially beneficial as additional informative nodes are strategically selected by the RL agent.

Interestingly, among the traditional ML and DL models, CNN and Gradient Boosting achieve relatively better performance compared to simpler models like Decision Trees and Random Forests, yet still significantly lag behind the Hybrid GNN. This further indicates the critical role of explicitly modeling network structure, something that general-purpose ML or DL methods inherently lack. 
In conclusion, explicitly modeling the network structure using Hybrid-GNN significantly surpasses traditional ML and DL methods, highlighting the importance of leveraging spatial relationships within bicycling networks and underscoring the effectiveness of our combined graph and attention based architecture. 

\subsection{Ablation Study} \label{sec:ablation_study}
To better understand the relative contributions of the individual components within our proposed \networkName{} framework, we conducted a comprehensive ablation study. Table~\ref{tab:ablation_study_table} compares the performance of the full \networkName{} (Hybrid GAT-GCN model combined with RL-driven sensor placement using Curiosity-Driven Exploration) against three ablated variants:

\begin{itemize} \item \textbf{Without RL:} The Hybrid GNN model was trained with additional sensors placed randomly, without any reinforcement learning guidance, to isolate the benefits specifically attributed to the RL-driven sensor placement strategy. 
    \item \textbf{Using Only GCN:} The Hybrid-GNN model was replaced by a simpler architecture containing only GCN layers (without GAT layers), maintaining the RL sensor selection via Curiosity-Driven Exploration. 
    \item \textbf{Using Only GAT:} Similarly, the Hybrid architecture was simplified to include only GAT layers (excluding GCN layers), again with RL sensor placement via Curiosity-Driven Exploration. 
\end{itemize}

\begin{figure*}
    \centering
    \includegraphics[width=1\linewidth]{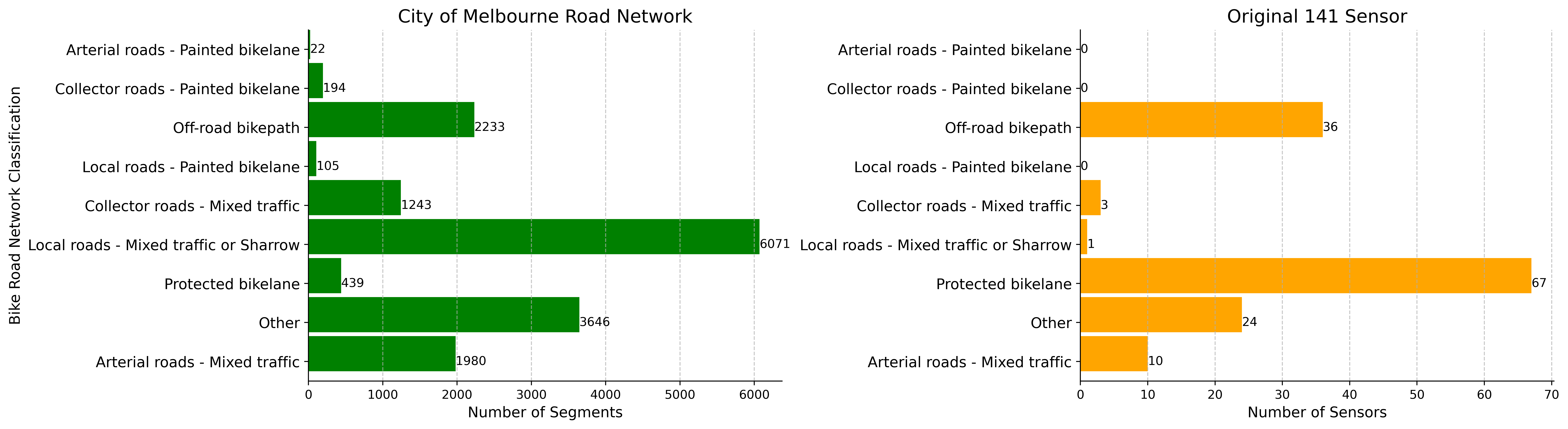}
    \caption{Comparison of road‐segment classification and existing sensor deployment in Melbourne’s bicycling network.     
    \emph{Left:} Total number of road segments by classification ( arterial roads, collector roads, local roads, off‑road bikeways, protected bikeways, mixed‑traffic segments, and others). 
    \emph{Right:} Distribution of the original 141 sensor locations across the same road classifications, highlighting current coverage imbalances.}

    \label{fig:melbourne road network vs original 141}
\end{figure*}

\begin{figure*}[t]
    \centering
    \includegraphics[width=1\linewidth]{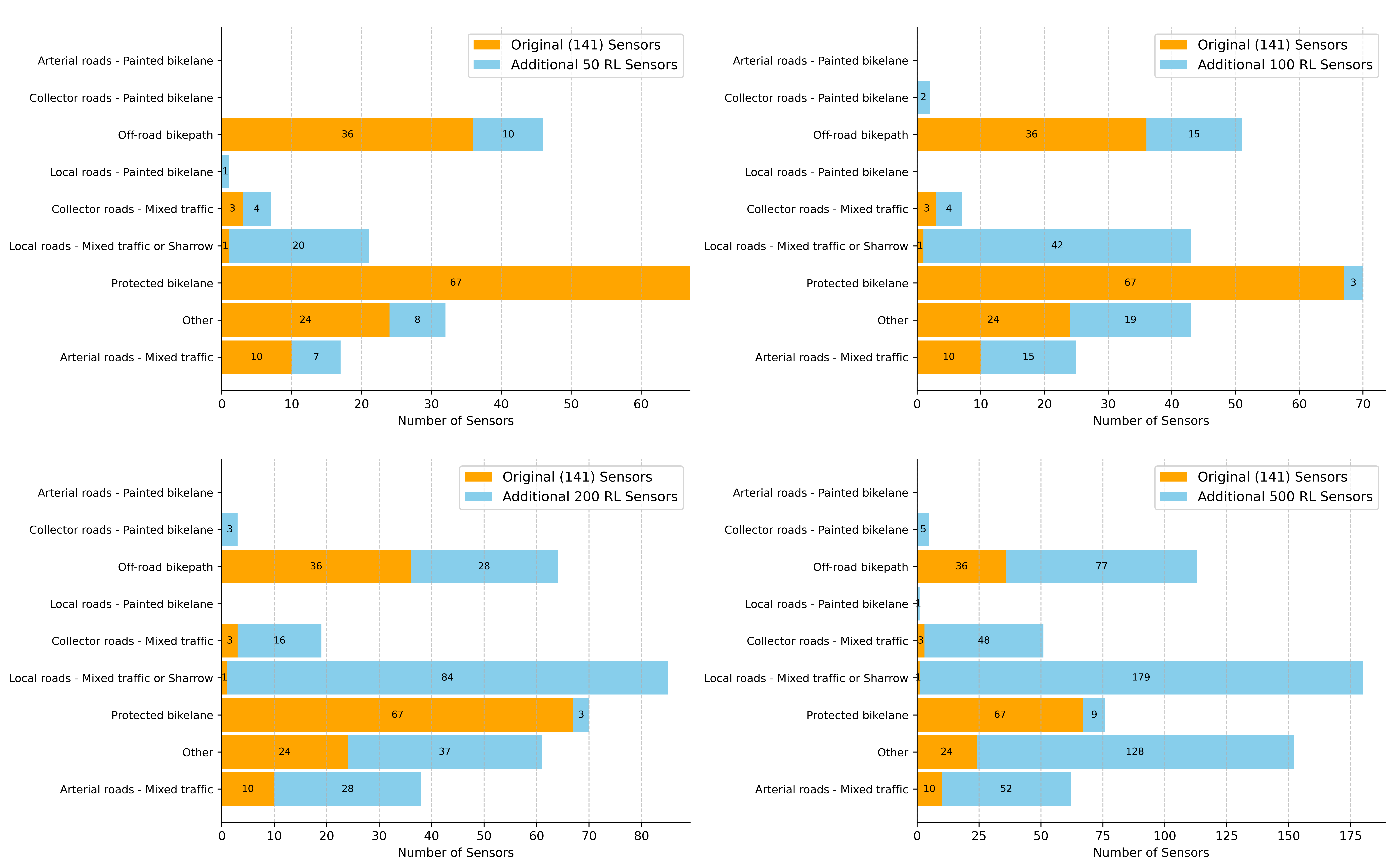}
    \caption{Distribution of original and RL‑selected sensor deployments across road classifications in Melbourne’s bicycling network. Each panel compares the count of original sensors (orange) with additional sensors chosen by the RL‑based Curiosity‑Driven Exploration policy (blue) for incremental deployments of 50 (top‑left), 100 (top‑right), 200 (bottom‑left), and 500 (bottom‑right) sensors across the road segment categories.}

    \label{fig:melbourne_sensor_placement_comparison_with_RL}
\end{figure*}

As shown in Fig.~\ref{fig:ablation_study_figure}, the full INSPIRE-GNN framework significantly outperformed all ablated variants across all metrics, highlighting the value of each integrated component. Notably, when reinforcement learning-driven sensor placement was replaced by random selection (i.e., removing the RL module), the model performance degraded sharply, with MSE increasing from 1020.48 to 2217.36. 
This substantial performance drop underscores the critical role of strategic, RL-based sensor placement in achieving accurate predictions.

Furthermore, the results demonstrate the complementary strengths of combining GCN and GAT layers. 
Models employing only GCN or only GAT layers, even with RL-driven sensor placement, performed worse than the Hybrid model. Specifically, the GCN-only variant exhibited an MSE of 1428.62, and the GAT-only variant achieved 1293.51, both substantially higher than the 1020.48 MSE obtained by the Hybrid-GNN. 
This indicates that capturing both global (via GCN) and localized relational dynamics (via GAT) simultaneously significantly enhances the accuracy of link-level bicycle volume estimation.

Overall, the ablation study validates the design choices behind the proposed INSPIRE-GNN architecture, demonstrating that both the hybrid graph architecture and the reinforcement learning-based sensor placement strategy are crucial for achieving optimal predictive performance in data-sparse bicycling networks.

\subsection{Implications of RL-Based Sensor Placement for Urban Bicycling Networks} \label{sec: implication}

Our analysis of Reinforcement Learning (RL)-based sensor placements in Melbourne's bicycling network provides valuable insights for transportation planners aiming to strategically expand sensor coverage in a cost-effective manner. 
By analyzing sensor deployment patterns, we identified how RL addresses critical gaps in existing sensor coverage and significantly improves the coverage and representativeness of collected data, leading to enhanced accuracy for bicycle volume estimation.

As shown in Fig.~\ref{fig:melbourne road network vs original 141}, initially Melbourne’s bicycling network had sensor coverage predominantly on protected bike lanes, while major gaps existed on other road classifications, particularly arterial roads with mixed traffic conditions and local roads. 
Specifically, of the initial 141 counters, nearly half (67) were placed on protected bike lanes, indicating a strong initial bias toward this infrastructure type. 
In contrast, arterial roads, local roads, and collector roads with mixed traffic conditions were substantially underrepresented.

The RL-based sensor placement strategy effectively mitigated these coverage gaps by strategically allocating additional sensors to previously neglected but significant segments as shown in Fig.~\ref{fig:melbourne_sensor_placement_comparison_with_RL}. Notably:

\begin{itemize}
    \item \textbf{Arterial Roads (Mixed Traffic):} Initially covered by only 10 sensors out of 1,980 segments, RL placements significantly increased sensor coverage by deploying 52 sensors among the additional 500 placements. 

    \item \textbf{Local Roads (Mixed Traffic or Sharrow):} Initially represented by only a single sensor, the RL approach allocated 20 out of 50 initial additional sensors to these segments, demonstrating a clear focus on filling critical gaps in quieter but potentially important cycling routes.
    
    \item \textbf{Off-road Bike Paths:} Initially somewhat well-represented, these paths received additional sensor deployments (up to 77 sensors in total), indicating a continued recognition of their significance in capturing broader recreational and commuting cycling activities.
\end{itemize}

Interestingly, the RL strategy consciously avoided adding more sensors to already adequately represented segments (such as protected bike lanes), prioritizing those road categories with significant initial data sparsity. 
This targeted approach substantially enhances the network's overall representativeness and maximizes the effectiveness of sensor deployments.

Traditionally, bicycling sensor placement decisions are often guided by the need to evaluate newly implemented infrastructure, such as protected bike lanes or dedicated paths, resulting in higher sensor density in those areas. This infrastructure-focused placement approach remains essential to directly measure the effectiveness of specific interventions. However, it may inadvertently neglect broader network coverage, leaving many segments underrepresented. The proposed INSPIRE-GNN framework complements this existing practice by providing a rigorous, data-driven method to strategically expand sensor networks, optimizing placements to achieve the greatest overall predictive accuracy. By balancing evaluation-focused placements with strategic expansions informed by reinforcement learning, cities can significantly enhance the reliability and comprehensiveness of their bicycling data collection efforts.

These findings underscore how employing an RL-driven sensor placement strategy can significantly enhance urban bicycling data collection, offering actionable insights for targeted infrastructure investments. 
By addressing under-monitored yet impactful segments, planners can leverage INSPIRE-GNN to improve infrastructure planning decisions, ultimately contributing to more sustainable and evidence-based urban mobility strategies.

\section{Conclusion and Future Directions}

This study introduced the \networkName{} framework, a novel hybrid Graph Neural Network (GNN) integrated with a Reinforcement Learning (RL)-based sensor placement strategy, specifically designed to enhance link-level bicycling volume estimation in urban networks characterized by high data sparsity. 
By combining Graph Convolutional Networks (GCN) and Graph Attention Networks (GAT) with a Deep Q-Network (DQN)-based RL framework, \networkName{} effectively addresses the critical issue of strategically placing additional sensors to maximize predictive performance. 
The empirical results demonstrate that the hybrid GNN architecture consistently outperforms traditional machine learning and deep learning models, including Decision Trees, Random Forests, Gradient Boosting, MLP, and CNN, across various scenarios of sensor deployment. 
Moreover, the proposed RL-based exploration approach significantly surpassed heuristic sensor placement strategies, including betweenness centrality, closeness centrality, random selection, and activity-based placement derived from Strava data. 
An ablation study provided deeper insight into the framework's effectiveness, highlighting the essential roles of both the GCN and GAT components, as well as the RL-based sensor selection strategy. 
The results clearly demonstrate that the combination of both graph convolutional and attention mechanisms substantially enhances predictive performance, underscoring the importance of capturing complex spatial dependencies within bicycling networks. 
Furthermore, the strategic selection of sensors using RL considerably improved accuracy by filling critical gaps left by initial sensor placements, thus providing robust estimations even under sparse conditions.

While our study offers significant insights, several promising avenues exist to extend and build upon this research. 
Further improvements can explore data integration and fusion approaches, combining Strava data with complementary datasets such as manual bicycle counts, GPS trajectories, or smartphone-based mobility data. Such integrated datasets would help mitigate potential biases associated with single source data, improve representativeness of cycling behavior across different cyclist groups and offer richer, more representative inputs for accurate bicycle volume modeling.
Additionally, by incorporating temporal dynamics explicitly, the Hybrid-GNN architecture can be expanded to capture variations across days, weeks, or seasons. This can yield further predictive accuracy improvements and inform dynamic sensor placement strategies. 
Future studies can also enhance the RL-based sensor placement strategy by incorporating multi-objective optimization, explicitly integrating practical considerations such as sensor deployment costs, maintenance complexity, installation feasibility, and equitable coverage across diverse cyclist demographics directly into the RL reward function. 
Lastly, evaluating our framework on more extensive sensor deployment scenarios or transferring the methodology to different urban contexts can also provide detailed insights into scalability, robustness, and transferability of this approach, facilitating its broader adoption in diverse urban planning scenarios to create safer, better-connected, and more sustainable bicycling infrastructure networks.

\subsection*{Acknowledgements}
The CYCLED (CitY-wide biCycLing Exposure modelling) Study is funded by an Australian Research Council Discovery Project (DP210102089).
Mohit Gupta's PhD scholarship and Debjit Bhowmick were supported by the Australian Research Council Discovery Project (DP210102089).
Ben Beck was supported by an Australian Research Council Future Fellowship (FT210100183). This work includes aggregated and de-identified data from Strava Metro \citep{StravaMetro}.
We would like to thank Dr Dana Kulic, Professor, Monash university, Australia for her guidance on the application of Reinforcement Learning in our study.

\bibliographystyle{apalike}
\bibliography{main.bib}
\clearpage
\appendix
\clearpage 

\end{document}